\documentclass[conference]{IEEEtran}
\usepackage{cite}
\usepackage{amsmath,amssymb,amsfonts}
\usepackage{algorithmic}
\usepackage{graphicx}
\usepackage{textcomp}
\def\BibTeX{{\rm B\kern-.05em{\sc i\kern-.025em b}\kern-.08em
    T\kern-.1667em\lower.7ex\hbox{E}\kern-.125emX}}
\begin{document}

\title{Mixup-Based Acoustic Scene Classification Using \\
Multi-Channel Convolutional Neural Network}

\author{\IEEEauthorblockN{Kele Xu}
	\IEEEauthorblockA{\textit{School of Information Communication} \\
		\textit{National University of Defense Technology}\\
		Wuhan, China \\
		kelele.xu@Gmail.com}
	
	\and
	\IEEEauthorblockN{Dawei Feng, Haibo Mi, Boqing Zhu}
	\IEEEauthorblockA{\textit{School of Computer} \\
		\textit{National University of Defense Technology}\\
		Changsha, China \\
		davyfeng.c@gmail.com, haibo\_mihb@126.com}
	
	\and
	\IEEEauthorblockN{Dezhi Wang, Lilun Zhang}
	\IEEEauthorblockA{\textit{College of Meteorology and Oceanography} \\
		\textit{National University of Defense Technology}\\
		Changsha, China \\
		wang\_dezhi@hotmail.com, zll0434@163.com}
	
	\and
	\IEEEauthorblockN{Hengxing Cai}
	\IEEEauthorblockA{\textit{School of Engineering} \\
		\textit{Sun Yat-Sen University}\\
		Guangzhou, China \\
		caihx3@mail2.sysu.edu.cn}

	\and
	\IEEEauthorblockN{Shuwen Liu}
	\IEEEauthorblockA{\textit{School of Computer Science} \\
		\textit{Nanjing University of Technology}\\
		Nanjing, China \\
		shuwenliu@njtech.edu.cn}
}

\maketitle

\begin{abstract}
Audio scene classification, the problem of predicting class labels of audio scenes, has drawn lots of attention during the last several years. However, it remains challenging and falls short of accuracy and efficiency. Recently, Convolutional Neural Network (CNN)-based methods have achieved better performance with comparison to the traditional methods. Nevertheless, conventional single channel CNN may fail to consider the fact that additional cues may be embedded in the multi-channel recordings. In this paper, we explore the use of Multi-channel CNN for the classification task, which aims to extract features from different channels in an end-to-end manner. We conduct the evaluation compared with the conventional CNN and traditional Gaussian Mixture Model-based methods. Moreover, to improve the classification accuracy further, this paper explores the using of mixup method. In brief, mixup trains the neural network on linear combinations of pairs of the representation of audio scene examples and their labels. By employing the mixup approach for data augmentation, the novel model can provide higher prediction accuracy and robustness in contrast with previous models, while the generalization error can also be reduced on the evaluation data.
\end{abstract}


\section{Introduction}

Acoustic scene classification (ASC) refers to the identification of the environment in which the audios have been acquired, which associates a semantic label to each audio. In 1997, Sawhney proposed the first method to address the ASC problem in an MIT technical report \cite{b1}. A set of classes, including ``people'', ``voices'', ``subway'', ``traffic'' is recorded. An overall classification accuracy of 68\% was obtained based on the recurrent neural networks and the K-nearest neighbor criterion. Indeed, the recognition of environments has become an important application in the field of machine listening, and ASC enables devices to make sense of their environments. The potential applications of ASC seem evident in several fields, such as security surveillance and context-aware services.

In order to solve the problem of lacking common benchmarking datasets, the first Detection and Classification of Acoustic Scenes and Events (DCASE) 2013 challenge \cite{b2} was organized by the IEEE Audio and Acoustic Signal Processing (AASP) Technical Committee. Many audio processing techniques have been proposed during the past years. The applications of deep learning in the ASC have witnessed a dramatic increase during last five years, especially the convolutional neural network (CNN). Compared to the traditional method, which commonly involves training a Gaussian Mixture Model (GMM) on the frame-level features such as Mel-Frequency Cepstral Coefficients (MFCCs) \cite{b3}, CNN-based methods can achieve better performance. However, most of the previous attempts aimed to apply the deep learning method by using one single channel (or just the average between the left and right channels) \cite{b4}. A robust Audio Scene classification model should be able to capture temporal patterns at different channels as additional cues may be embedded in the multi-channel recordings \cite{b5}. In this paper, we explore the use of multi-channel CNN for the ASC task, which achieves better accuracy with comparison to the standard CNN.

On the other hand, the deep neural network architectures have a large number of parameters, and they are prone to overfitting. The easiest and most widely used approach to reduce overfitting is to employ larger datasets. As an alternative, data augmentation method can be used to improve the performance of neural network by artificially enlarging the dataset using label-preserving transformation. However, only a few attempts have been made for the data augmentation for audio scene classification.

In this paper, we explore the use of mixup-based method for data augmentation \cite{b6}, with the goal to obtain superior accuracy and robustness. In brief, mixup constructs virtual training examples, and the neural network can be trained by using the linear combinations of pairs of the representation of examples and their labels.

Theoretically, mixup extends the training distribution by incorporating the prior knowledge that linear interpolations of audio feature vectors should lead to linear interpolations of the associated targets \cite{b6}. Mixup can be implemented in a few lines of code, and induces the minimal computation overhead. Despite its simplicity, mixup allows a performance improvement using the DCASE 2017 audio scene classification dataset.

The paper is organized as follows. Section 2 discusses the relationship between our method and prior work, while, the multi-channel CNN classification method is presented in Section 3. Section 4 describes the mixup method, and the experimental results are given in Section 5. Section 6 gives the conclusion of this paper.

\section{Related to Prior Work}

Scene classification (detection) has been explored by computer vision using different techniques, and dramatic progress has been made during last two decades. However, compared to the progress of scene classification using image (or video), audio-based approaches have been under-explored, and the state-of-the-art audio-based techniques are not able to achieve the comparable performance to its image/video counterpart. In fact, audios can sometimes be more descriptive than videos/images, especially when it comes to the description of an event.

Recently, due to the release of the relatively larger labeled data, there has been a plethora of efforts have been made for the audio scene classification task \cite{b7}, \cite{b8}. In brief, the main contributions can be divided into three parts: the representation of the audio signal (or handcrafted feature design) \cite{b9},  \cite{b10}, \cite{b11}; more sophisticated shallow-architecture classifiers \cite{b12}, \cite{b13}, \cite{b14} and the applications of deep learning in ASC task \cite{b15}, \cite{b16}. 

Indeed, deep learning has witnessed dramatic progress during the last decade and achieved success in several different fields, such as, image classification \cite{b16}, speech recognition \cite{b17}, natural language processing \cite{b18} and so on. Although, there are some attempts, which employ CNN as the tool to solve the ASC task, most of them tried to solve the problem within the context of using the monaural signals. In \cite{b11}, the author proposed to concatenate different channels, resulting in a one-channel file with longer duration. This kind of method employed the one-channel CNN architecture. In \cite{b20}, the author proposed to use all-convolutional neural network and masked global pooling for the ASC task. However, only left-hand channel was employed for the classification task. Here, we argue that additional cues may be embedded in the binaural recordings \cite{b11}. The combination of information in multi-channels may lead to advanced feature representations for the classification.

On the other hand, the trend of deep neural network' architecture is to become deeper and wider, and millions of parameters need to be trained. To improve the generalization ability of neural networks, plenty of regularization approaches have been used, which include: batch normalization, dropout, etc. When there is only limited training data available, data augmentation using preserving transformation is a widely-used technique for the neural network training to improve the robustness. Although following the same concept of improving the prediction invariance of deep neural network, the data augmentation in audio scene classification is different from the image classification tasks, and the traditional augmentation, such as rotation, flipping, distorting and deformation cannot be applied directly. The procedure is dataset-dependent and requires the use of expert knowledge \cite{b6}.

In this paper, we explore the use of mixup data augmentation approach, which was proposed in \cite{b6}. In brief, the new samples are created by mixing two inputs of the neural network with a ratio, and the labels of the samples are similar to the between-class label. Normally, the ratio ranges from 0 to 1. Using the DCASE 2017 audio scene classification dataset, improved performance has been observed after employing mixup approach. 

\section{Multi Channel Convolutional Neural Network}
Due to its ability of automatic learning complex feature representations, CNNs have achieved great success. CNN has the potential to identify the various salient patterns of the audio signals. In more detail, the processing units in the lower layers can obtain the local feature of the signals, while the higher layers can extract the features of high-level representations.

The input for a CNN architecture can be the raw audio signal or the spatial frequency representation of the raw signal (for example: MFCCs, Short time Fourier transform, spectrograms). In our experiments, we employ the widely used feature representation: Mel-filter bank features of the audio signal segments as the input for the CNN. However, it is not complicated to extend our framework for other kinds of input.

Unlike the attempts which aim to maintain the one-channel CNN architecture \cite{b11}, we extract features in terms of three different channels. The three different channels are: left channel, right channel, the mean between the left and right channels. The Mel-filter bank features of different channels will be concatenated as a multi-channel image, which results in training a system in an end-to-end manner. Note that, the Mel-filter bank features configuration was kept the same for each single channel during our experiments. In our experiment, Mel-filter bank features is calculated for each channel. We employ the first half of the symmetric Hann window as the window function with a window size of 25ms and a hop size of 25ms.

The input of the network is three-channels Mel-filter bank features with size 3$\times$128$\times$128, where 3 represents the number of channels, 128$\times$128 denotes the size of Mel-filter bank features for single channel. The input sizes are kept the same during the experiments. The flowchart of Multi-channel CNN-based audio scene classification is given in Fig 1.

\begin{figure}[!ht]
	\centering
	\includegraphics[width=80mm]{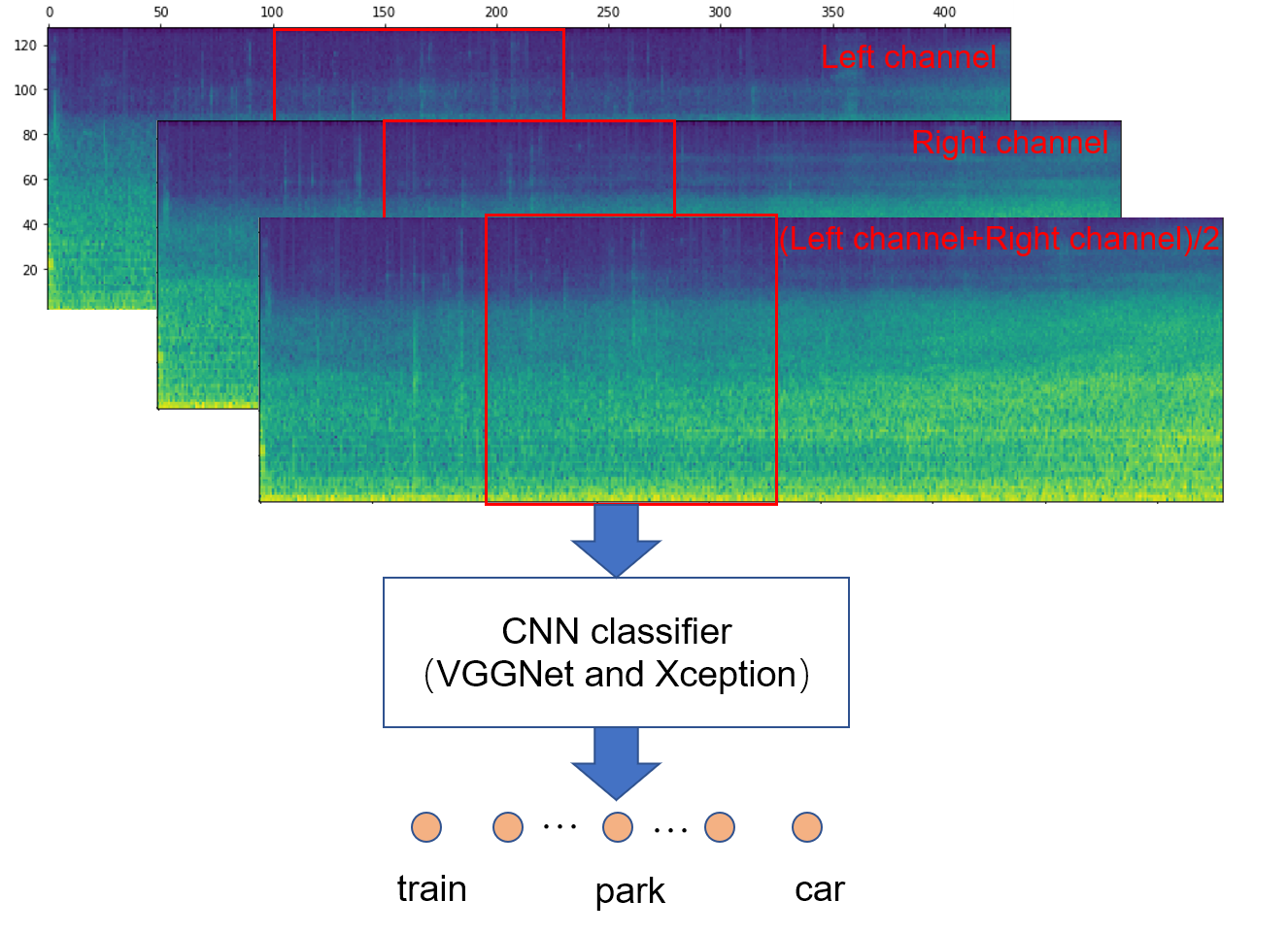}
	\caption{Multi-channel CNN-based audio scene classification}
	\label{fig1}
\end{figure}

There are numerous variants of CNN architectures in the literature. However, their basic components are very similar. Since the starting with LeNet-5 \cite{b23}, convolutional neural networks have typically standard structure-stacked convolutional layers (optionally followed by batch normalization and max-pooling) are followed by fully-connected layers.

In this paper, we followed the VGG-style \cite{b24} networks and Xception \cite{b25} networks due to its relatively high accuracy and simplicity. The main contribution of VGG net is to increase the depth using an architecture with very small (3$\times$3) convolution filters.

While VGG achieves an impressive accuracy on the image classification task, its deployment on even the most modest-sized GPUs is a problem because of huge computational requirements, both in terms of memory and time. It becomes inefficient due to large width of convolutional layers.

As the-state-of-the-art model in Inception model group, Xception architecture employs the depthwise separable convolution operation to replace the regular Inception modules, which has an excellent performance on a larger image classification dataset like ImageNet, and becomes a cornerstone of convolutional neural network architecture design. Another change that Xception model made, was to replace the fully-connected layers at the end with a simple global average pooling which averages out the channel values across the 2D feature map, after the last convolutional layer. This drastically reduces the total number of parameters. This can be understood from VGGNet, where fully connected layers contain about 90\% of parameters.

The only changes we made to VGG were to the final layer (using the global average layer) as well as the use of batch normalization instead of Local Response Normalization (LRN). The parameters of the CNN model are optimized with stochastic gradient descent. The cross-entropy was selected as the objective function. Moreover, an L2 weight decay penalty of 0.002 was employed in our model. To train the CNN, we used Keras with tensorflow backend, which can fully utilize GPU resource. CUDA and cuDNN were also used to accelerate the system.

It is worthwhile to note that each layer consists of many convolutions or pooling operators. The convolutional filters can be interpreted as the filter-banks learning. For the activation layer, the rectified linear unit is used to introduce the non-linearity into a neural network. The last layer is the probability output layer, that converts the output vector of the fully connected layer to a vector of probabilities, which sum up to 1, each probability corresponding to one class. The probabilities can be used to predict the scene label of the audio segment.

For the final prediction of an input instance, there are many widely used approaches to perform the final prediction, for example, maximum probability, median probability, average probability and majority votes. In this paper, for the evaluation of the CNN-based method, we use the maximum probability to obtain the label.
 
\section{Mixup For Data Augmentation}

We evaluate the multi-channel CNN on the TUT sound events detection 2017 database \cite{b7}. The database consists of stereo recordings which were collected using 44.1 kHz sampling rate and 24-bit resolution. The recordings came from 15 various acoustic scenes, which have distinct recording locations, for example: office, train, forest path. For different locations, 3-5 minutes long audio was recorded. And the audio files were split into 30-second segments. The acoustic scene classes considered in this task were: bus, cafe/restaurant, car, city center, forest path, grocery store, home, lakeside beach, library, metro station, office, residential area, train, tram, and urban park.

Currently, most publicly available ASC datasets have limited sizes \cite{b3}, \cite{b7}. The disadvantage of small dataset is that the model is prone to overfitting. In the DCASE 2017 audio scene classification task, it is found that the generalization gap is big, and the accuracy difference between development dataset and evaluation dataset ranges from 4\% to 30\% by using different approaches. The ability to generalization is a research topic for the deep neural network. To improve the generalization ability of deep neural network, especially the CNN, a plethora approaches have been proposed, such as dropout \cite{b26}, batch normalization\cite{b27}. Data augmentation is another explicit form of regularization, which is also widely used in the deep neural network. In more detail, for the deep CNN, random cropping and random flipping are two most popular data augmentation approaches. However, these methods cannot be applied to ASC directly. Recently, it is found that Generative adversarial network can be used for ASC data augmentation, and impressive performance have been obtained on the task \cite{b28}. Indeed, the data augmentation is under-explored in previous ASC study.

In this paper, we explore the use of mixup data augmentation. In more detail, virtual training examples can be constructed by using the following formula:
\begin{equation}
x=\alpha \times x_i + (1-\alpha) \times x_j
\end{equation}
\begin{equation}
y=\alpha \times y_i + (1-\alpha) \times y_j
\end{equation}

Where ($x_i$, $y_i$) and ($x_j$, $y_j$) are two examples random selected from the training data of the DCASE 2017 ASC task. $\alpha$ is the mixed ratio. In our experiments, $\alpha \in [0,1]$. A mixup example is given in Fig 2, and the $\alpha$ is set as 0.2 for the example. In the figure, two labeled audio scenes are selected randomly, while a new training sample is constructed by weighted average between two given samples.

\begin{figure}[!ht]
	\centering
	\includegraphics[width=80mm]{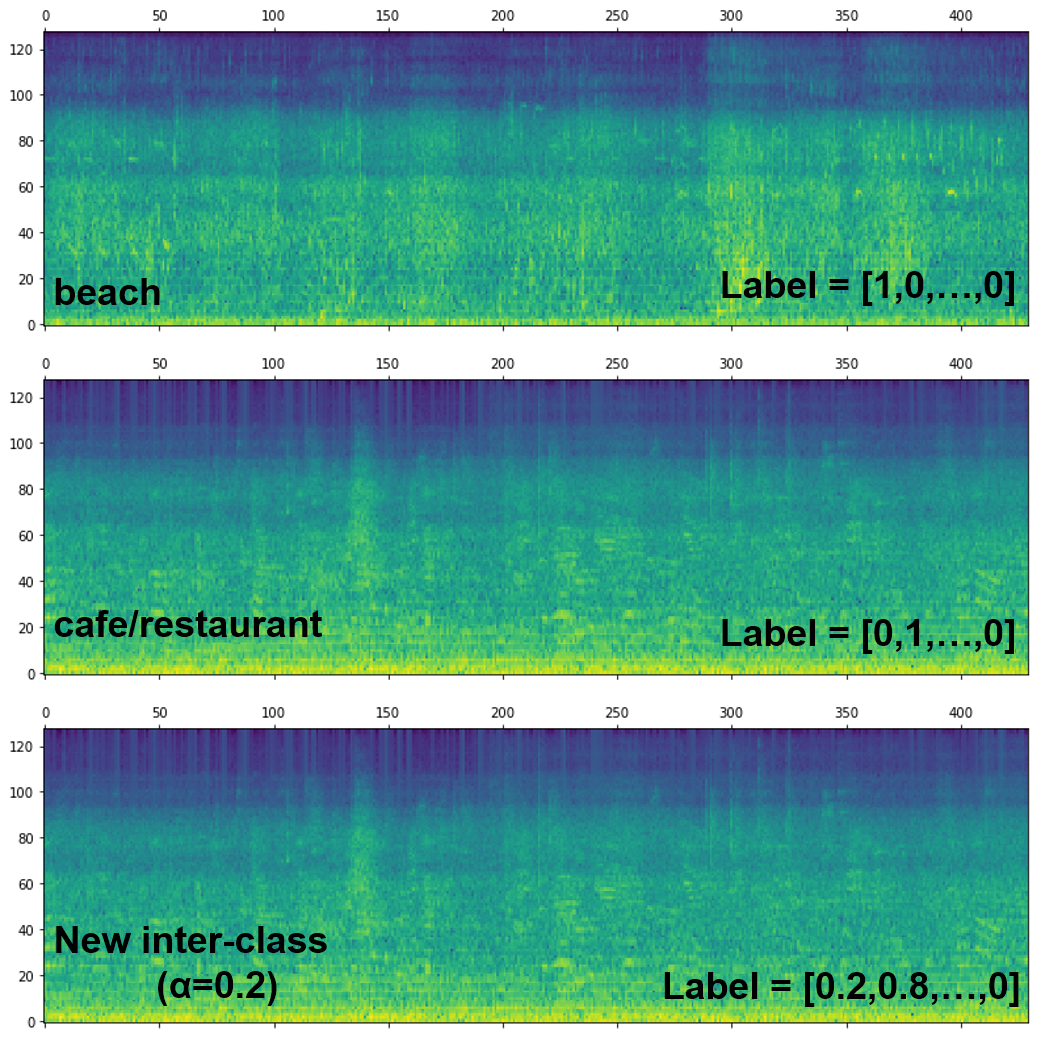}
	\caption{An example for mixup data augmentation for audio scene classification.}
	\label{fig2}
\end{figure}

Despite its simplicity, the mixup data augmentation methods have provided state-of-the-art performance in many datasets, which include the CIFAR-10, CIFAR-100, and ImageNet-2012 image classification datasets. Similar to create inter-class, mixup increases the robustness of deep CNN when the samples contains corrupted labeled ones. In the following section, we will demonstrate that mixup data augmentation can also improve the ASC performance.

\section{Experimental Results}\label{ER}

As the DCASE 2017 scene classification dataset provides cross-validation splits, we follow the 4 fold cross-validation splits. We used the same experiment settings from development set for the evaluation set. In the development stage, the results are evaluated in terms of average accuracy for 4 folds. The performance of evaluation data is also given in this section.

In our experiments, we made two sets of comparison: the performance comparison between the single channel CNN and multi-channel CNN; the performance comparison between Multi-channel CNN with mixup data augmentation and Multi-channel CNN without mixup data augmentation.

\subsection{Single/Multi-channel CNN}

The first set of experience aims to evaluate single-channel and multi-channel based audio scene classification using convolutonal neural network. As aforementioned, the architectures used for the comparison include: VGGNet and Xception, and all of the CNNs are trained from scratch without any pre-trained initialization. Table 1 presents the validation results for the 4 fold cross-validation as well as the performance on the evaluation data. The performance of baseline is also given in Table 1. In more detail, the baseline system used here consists of 60 MFCC features and a Gaussian mixture model (GMM) based classifier.
\begin{table}[htbp]
	\caption{Audio scene classification accuracy using single/multi channel Convolutional Neural Network}
	\begin{center}
		\begin{tabular}{|c|c|c|}
			\cline{1-3} 
			\textbf{\textit{Method}}& \textbf{\textit{Cross-validation}}& \textbf{\textit{Evaluation}} \\
			\hline
			Baseline (GMM) & 74.8\%  & 61.0\%  \\
			Single-channel using VGGNet & 82.4\%  & 67.8\%  \\
			Multi-channel using VGGNet & 84.7\%  & 71.5\%  \\
			Single-channel using Xception & 83.3\%  & 72.1\%  \\
			Multi-channel using Xception & 85.4\%  & 74.5\%  \\
			\hline
		\end{tabular}
		\label{tab1}
	\end{center}
\end{table}
As can be seen from the table, both single-channel CNNs and multi-channel CNN showed better performance against the baseline. It can also be observed that multi-channel CNN performance is better than the single-channel CNN using different architectures, and the accuracy is increased about 2\% to 3\%. It may imply that additional features can be extracted from multi-channels, which can improve the accuracy for the ASC task. Similar to the results obtained on the ImageNet classification, Xception architecture provides better performance than VGGNet. The reason is that Xception architecture can take multi-scale information into account as different kernels size are learned in the model.
On the other hand, the accuracy difference between the development dataset and evaluation dataset) is significant, which ranges from 10.9\% to 14.2\% using different approaches. This may indicate that the trained CNN models are prone to overfitting during the training procedure.

\subsection{Multi-channel CNN with/without mixup}

The second set of experiments aimed to show that mixup-base data augmentation can improve the performance of ASC. Moreover, it is also shown that mixup can reduce the generalization gap.

\begin{table}[htbp]
	\caption{Audio scene classification accuracy using mixup data augmentation}
	\begin{center}
		\begin{tabular}{|c|c|c|c|}
			\cline{1-4} 
			\textbf{\textit{Method}} & \textit{$\alpha$}&  \textbf{\textit{Cross-validation}}& \textbf{\textit{Evaluation}} \\
			\hline
			Multi-channel VGGNet & 0   & 84.7\%  & 71.5\%  \\
			Multi-channel VGGNet & 0.2 & 85.2\%  & 73.4\%  \\
			Multi-channel VGGNet & 0.5 & 86.9\%  & 73.2\%  \\
			Multi-channel VGGNet & 0.8 & 85.8\%  & 72.1\%  \\
			Multi-channel Xception & 0 & 85.4\%  & 74.5\%  \\
			Multi-channel Xception & 0.2 & 86.7\%  & 75.6\%  \\
			Multi-channel Xception & 0.5 & 87.2\%  & 76.7\%  \\
			Multi-channel Xception & 0.8 & 86.9\%  & 74.8\%  \\
			\hline
		\end{tabular}
		\label{tab2}
	\end{center}
\end{table}

The performance of different approaches, which employing the mixup data augmentation, is given in Table 2, and different ratios are used for the mixup. Due to the computation resource constraint, only three ratios are used in our experiments (when $\alpha = 0$, the mixup approach is not employed). As can be seen from the table: without mixup data augmentation, the cross-validation accuracy of multi-channel VGGNet is 84.7\% and the accuracy of evaluation data is 71.5\%, while the accuracy of evaluation data ranges from 72.1\% to 73.2\% if mixup data augmentation is employed. For multi-channel CNN using Xception architecture, the accuracy was also improved using the mixup data augmentation, which demonstrates that mixup approach is effective despite its simplicity. In our experimental results, mixup with ratio 0.5 provides superior performance.

\section{Conclusion}
In this paper, we have presented the multi-channel convolutional neural network-based method for the multi-class acoustic scene classification.
To summarize, the contributions of this paper are twofold: firstly, we present a multi-channel CNN architecture for the classification task. Secondly, we explore the mixup data augmentation method, and experiments demonstrated that by employing the mixup dataset augmentation, the classification can be improved, and the generalization error can also be reduced. To the best knowledge of the authors, this is the first attempt of employing mixup for the audio scene classification task.
For future work, we will investigate the CNN architecture to utilize multi-scale information embedded in the audio signal, thus improving the classification accuracy. The mixup approach is also needed to be fully explored. Presently, the mixup processing is relied on the log mel spectrum of the audio signal. We did not observe significant  improvement by mixup of the raw audio signal directly. Moreover, mixup may also be useful for audio event tagging and detection.

\section*{Acknowledgment}

We would also like to thank the Qiuqiang Kong from the CVSSP group (Surrey University), for providing the constructive suggestions and help during the experiments.

\end{document}